\documentclass[letterpaper, 10 pt, conference]{ieeeconf}  

\IEEEoverridecommandlockouts                              

\usepackage{graphicx}
\usepackage{subcaption}
\usepackage{verbatim}
\usepackage{fancyvrb}
\usepackage{framed}
\usepackage[hidelinks]{hyperref}
\usepackage{xcolor}

\usepackage{color}

\newif\ifblind
\blindfalse
\definecolor{darkgreen}{rgb}{0,0.4,0}
\title{\LARGE \bf
HARMONIC: Cognitive and Control Collaboration in Human-Robotic Teams
}

\ifblind
\author{}

\else
\author{Sanjay Oruganti$^{1}$, Sergei Nirenburg$^{1}$, Marjorie McShane$^{1}$, Jesse English$^{1}$, Michael K. Roberts$^{1}$,\\ Christian Arndt$^{1}$, Sahithi Kamireddy$^{2}$, Carlos Gonzalez$^{3}$, Mingyo Seo$^{4}$, Luis Sentis$^{3}$
\thanks{$^{1}$Cognitive Science Department and $^{2}$Department of Computer Science, Rensselaer Polytechnic Institute, Troy, NY 12180, USA. Departments of $^{3}$Aerospace Engineering and Engineering Mechanics, and ${^4}$ Electrical and Computer Engineering, at the University of Texas at Austin, TX, US.
        {\tt\small e-mail: sanjayovs@ieee.org}}%
}
\fi

\begin{document}

\maketitle
\thispagestyle{empty}
\pagestyle{empty}

\begin{abstract}
This paper describes HARMONIC, a cognitive-robotic architecture that integrates the OntoAgent cognitive framework with general-purpose robot control systems applied to human-robot teaming (HRT). HARMONIC incorporates metacognition, meaningful natural language communication, and explainability capabilities required for developing mutual trust in HRT. Through simulation experiments involving a joint search task performed by a heterogeneous team of two HARMONIC-based robots and a human operator, we demonstrate heterogeneous robots that coordinate their actions, adapt to complex scenarios, and engage in natural human-robot communication. Evaluation results show that HARMONIC-based robots can reason about plans, goals, and team member attitudes while providing clear explanations for their decisions, which are essential requirements for realistic human-robot teaming.

\end{abstract}

\section{Introduction}





 Current state-of-the-art robotic systems demonstrate remarkable tactical proficiency, from the sub-millimeter precision of surgical robots performing surgical procedures \cite{rivero2023robotic} to the dynamic multi-terrain locomotion of quadrupedal \cite{wetzel2022use} and humanoid platforms exhibiting sophisticated whole-body coordination \cite{tong2024advancements}. These systems also excel at integrating multimodal sensory information through sensor fusion, simultaneously localizing themselves within complex environments, recognizing objects, tracking moving targets with increasing reliability, in adaptive policy learning and multi-robot planning and collaboration \cite{torreno2017cooperative, antonyshyn2023multiple}. 

Despite these significant advancements in tactical execution, most state-of-the-art robots are limited to short-horizon planning and execution and lack the strategic capabilities necessary for long-horizon planning, reasoning, and true collaborative partnership with humans in complex decision-making scenarios \cite{natarajan2023human}. Importantly, effective human-robot teaming \cite{sharma2023review} requires more than just tactical proficiency; to function as trusted partners within human-robot teams, they must exhibit a range of cognitive and metacognitive capabilities, including: 

\begin{itemize}
\item developing and refining an understanding of team tasks and organization;
\item building specific capabilities related to the responsibilities, preferences, and actions of self and other team members \cite{nirenburg2024mutual};
\item maintaining and using episodic memory to leverage past experiences for predicting future needs and actions;
\item interpreting language, visual, and haptic perception through an underlying formal model of the world (ontology) \cite{mcshane2024agents};
\item communicating with teammates using meaningful, interpreted natural language.
\end{itemize}


\begin{figure}
    \centering
    \includegraphics[width=\linewidth]{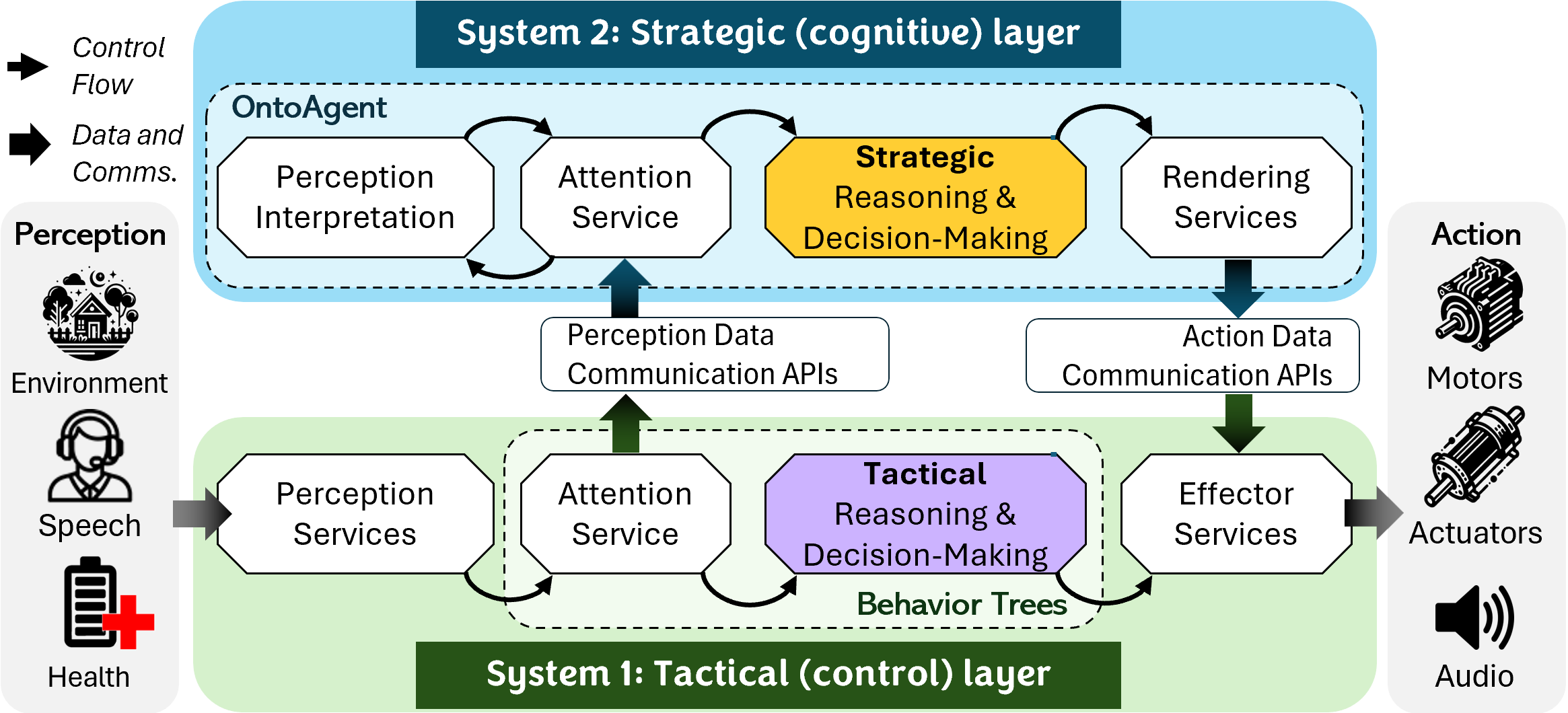}
    \caption{An overview of the HARMONIC framework, showing the Strategic and Tactical components representing the high-level planning (System 2) and low-level execution (System 1), respectively.}
    \label{fig:overview}
    \vspace{-18pt}
\end{figure}
Long-horizon functioning is within the scope of the field of cognitive systems, where much progress has been made on the tasks of complex agent team interactions, managing attention, goal selection, and plan execution. Still, while cognitive systems address key limitations in current robotic systems, they do not traditionally concentrate on the dynamic world of embodied agents. This has resulted in a disconnect between cognitive systems and practical robotic control systems. Our work integrates and leverages the strengths of both these areas of research and development. 

The HARMONIC framework uses a dual-layer architecture that integrates strategic cognitive capabilities with tactical robotic control. This integration not only enables long-horizon planning but also enhances the scope of the robot's reasoning capabilities, adds metacognitive awareness and social intelligence, which in turn supports human-level explanation that serves as a foundation for robots to function as trusted teammates \cite{nirenburg2024explaining}, rather than mere tools, within human-robot teams.

HARMONIC incorporates the cognitive architecture OntoAgent \cite{english2020ontoagent,mcshane2021linguistics,mcshane2024agents} as a strategic layer. Within  HARMONIC, OntoAgent is responsible for semantic interpretation of sensory perception, grounding, attention management, goal-setting, sophisticated planning (supporting diverse multi-agent planning strategies),  and addressing unexpected challenges in ways that humans can readily interpret. Additionally, OntoAgent supports meaning-oriented natural language communication, enabling clear causal explanations for the robots' actions and decisions \cite{nirenburg2024explaining}. This capability significantly enhances skill transfer and knowledge sharing within human-robot teams. The use of the OntoAgent framework greatly enhances transparency of decision-making in HARMONIC-based robots, establishing crucial foundations for trust in human-robot collaboration \cite{mcshane2024agents}.


We present our approach to human-robot teaming using the example of a team comprised of two robots and a human executing a joint task. We demonstrate the robots' knowledge of team organization as well as their ability to (a) dynamically negotiate tasks and allocate responsibilities, (b) select and modify plans, and (c) adapt language communication to both the situation and the needs of their interlocutors. We implement these robotic capabilities using the novel HARMONIC architecture that operates in parallel at two levels -- the strategic cognitive level, which supports operations requiring reasoning, and the tactical control level, which supports skill-based, ``automatic," reflexive operation of robots.

Specifically, the key contributions of this paper are as follows:

\begin{itemize}
    \item We introduce HARMONIC, a cognitive-robotic architecture, which enables a flexible integration of OntoAgent \cite{english2020ontoagent}, a cognitive architecture with robot planning and control using Behavior Trees \cite{colledanchise2018behavior}.
    \item A cognitive strategy for multi-robot planning and execution with metacognition, communication, and explainability. This includes the robots' natural language understanding and generation, their reasoning about plans, goals, and attitudes, and their ability to explain the reasons for their own and others' actions.
    \item Results of simulation experiments on a search task carried out by a robotic team, which showcases the human-robot team's interactions and ability to handle complex, scenarios.
\end{itemize}

 Section \ref{sec:background} briefly describes the state of the art and background. Section \ref{sec:harmonicSec} introduces the HARMONIC framework and the implementation of its strategic and tactical levels. 
 Section \ref{sec:plans} describes the task, the team configuration, and the development of distributed plans, and Section \ref{sec:evaluation} presents the evaluation.


\section{Background and Related Work}
\label{sec:background}
\subsection{Cognitive-Robotic Architectures}
There are relatively few cognitive architectures explicitly designed or effectively adapted for robotic applications. Among these architectures, Soar \cite{laird2012cognitive}, ACT-R \cite{ritter2019act}, and DIARC \cite{scheutz2013systematic} are prominent, each contributing distinct methods and philosophies toward cognitive autonomy in robotics.

The Soar architecture provides symbolic cognitive processing through rule-based decision-making, facilitating general problem-solving and reasoning \cite{laird2012cognitive}. It excels at clear, structured decision-making but requires extensive manual integration for perception and motor control, making it challenging to deploy in unpredictable or sensor-rich environments without substantial customization. The ACT-R architecture \cite{ritter2019act}, on the other hand, models human cognition through symbolic reasoning combined with sub-symbolic activations, supporting sophisticated simulations of human cognitive processes like learning and memory. Its psychological plausibility makes it ideal for human-robot interaction scenarios, such as storytelling or social robotics, yet its heavy reliance on detailed knowledge engineering restricts flexibility in complex and dynamic environments.


DIARC is explicitly designed for cognitive robotics with distributed modules \cite{scheutz2013systematic,schermerhorn2006diarc}, affective processes, and natural language capabilities suitable for dynamic human-robot interactions in narrow domains. However, it struggles with concurrent process management and lacks clear architectural layering, unified ontology, explainability, long-term episodic memory, explicit metacognition, and formalized safety mechanisms, thus limiting its robustness and adaptability. In summary, most architectures require extensive customization from both cognitive and control perspectives, leading to limited real-world applications and practical testing.

\subsection{Multi-Agent Planning (MAP) and Communication}

Multi-Agent Planning (MAP) in Heterogeneous Multi-Robot (HMR) teams includes a variety of centralized, decentralized, and hybrid strategies\cite{amato2015planning, torreno2017cooperative, le1990combination}, each presenting distinct trade-offs. None of the current MAP approaches incorporates sufficient means of intra-team communication, which is an essential requirement for robust HMR teaming. Such communication must involve distinct protocols for human-robot and robot-robot interactions \cite{Soon2019ALanguage}. A unified, human-centric communication approach within a common cognitive architecture is advocated by Lemon et al. \cite{lemon2022conversational} and Natarajan et al. \cite{natarajan2023human}. HARMONIC, with OntoAgent as its cognitive strategic layer, implements human-oriented communication, reasoning, and team dynamics.  It also incorporates a variety of these scientific advances, such as the notion of common ground introduced by Clark et al. \cite{clark1996using} and further developed by Klein et al. \cite{klein2005common}, the latest in dialogue processing systems \cite{allen2001architecture}, grounding techniques \cite{roy2005grounding,lindes2017grounding}, work on team organization \cite{nirenburg1986providing} and natural language communication \cite{mcshane2024agents}.

\subsection{The OntoAgent Cognitive Architecture}
\label{sec:OntoAgent}
OntoAgent is a cognitive architecture, designed to support the development of social intelligent agents supporting content-centric computational cognitive modeling \cite{mcshane2021linguistics,mcshane2024agents,english2020ontoagent}. This approach emphasizes the need to acquire, maintain, and dynamically expand large-scale knowledge bases, which are essential for an agent's perception interpretation, reasoning, and action. The architecture's memory structure is divided into three main components: a Situation Model (SM) that contains currently active concept instances; a Long-Term Semantic Memory (LTS) that stores knowledge about instances of events and objects, and an Episodic Memory (LTE) that stores knowledge about instances of events and objects. OntoAgent supports goal-directed behavior through a goal agenda and a prioritizer that selects which goals to pursue. It typically uses stored plans associated with goals but can also engage in reasoning from first principles when necessary.

The architecture is built on a service-based infrastructure, with key services including perception interpretation, attention management, reasoning, and action rendering. OntoAgent can integrate both native services and imported external capabilities, such as robotic vision or text-to-speech modules. This flexibility allows it to support multiple perception channels, including language understanding, visual perception, and simulated interoception (Chapter 8 of \cite{mcshane2021linguistics}).

A crucial component of OntoAgent is OntoGraph, a knowledge base API that provides a unified format for representing and accessing knowledge across the system. OntoGraph supports inheritance, flexible organization of knowledge into "spaces," and efficient querying and retrieval. It implements a graph database view of knowledge, allowing for complex relational queries and supporting the representation of inheritance hierarchies.

OntoAgent places a strong emphasis on natural language understanding (NLU) and meaning-based text generation. These capabilities rely on a semantic lexicon that links words and phrases with meanings grounded in a resident ontology. The language analyzer treats a large range of complex linguistic phenomena (lexical disambiguation, reference resolution, ellipsis reconstruction, new-word learning, etc. \cite{mcshane2024agents,mcshane2021linguistics}) and generates ontologically-grounded interpretations of text meaning. The language generator, for its part, generates natural language utterances from ontologically grounded meaning representations that the agent creates as part of its reasoning about action. OntoAgent also supports ontological interpretation of visual percepts using an opticon whose entries link images to ontological objects. 

The architecture supports goal-directed behavior through a goal agenda and a prioritizer that selects which goals to pursue. It typically uses stored plans associated with goals but can also engage in reasoning from first principles when necessary. In the current work, OntoAgent is incorporated in the strategic layer of HARMONIC.




%

\section{The HARMONIC Framework}
\label{sec:harmonicSec}

HARMONIC (Human-AI Robotic Team Member Operating with Natural Intelligence and Communication), shown in  Fig. \ref{fig:overview}, is a dual-control cognitive-robotic architecture that integrates strategic, cognitive-level decision-making with tactical, skill-level robot control. It builds upon and extends the concept of hybrid control systems and architectures, as discussed in a comprehensive review by Dennis et al. \cite{dennis2016practical}. It also represents an advance over the type 2 integration approach employed in DIARC \cite{scheutz2013systematic, schermerhorn2006diarc}. Unlike DIARC, where the strategic layer is embedded as a subsystem within the tactical layer to enable concurrent and dynamic operations, HARMONIC introduces a more sophisticated integration of these components.

The strategic and tactical architectural layers of HARMONIC are connected by a bidirectional interface for seamless communication and data transfer. The strategic layer implements the OntoAgent cognitive architecture \cite{english2020ontoagent, mcshane2021linguistics,mcshane2024agents}, encompasses modules for attention management, perception interpretation, and decision-making (see Sec. \ref{sec:OntoAgent}). It employs both utility-based and analogical reasoning, enhanced by metacognitive abilities. This layer prioritizes strategic goals, manages plan agendas, and selects actions while continuously monitoring their execution. It also facilitates team-oriented operations, including communicating in natural language and explaining decisions.

When the strategic layer receives perception inputs, including speech, vision, and robot state information, as data frames from the tactical layer, it interprets them in the context of its active goals and situation model. These inputs are transformed into ontologically grounded meaning representations, enabling unified reasoning across modalities. This interpretation supports downstream processes such as attention management and strategic decision-making by allowing the agent to evaluate the relevance of new information, update its understanding of the environment, monitor progress on current plans, and determine whether adjustments are necessary. Based on this reasoning, the OntoAgent in the strategic layer may issue updated action commands or generate natural language responses to teammates.

The tactical layer, on the other hand, is responsible for reflexive attention management, robot action planning and control, sensor input processing, and executing motor actions based on high-level commands received from the strategic layer. The attention manager processes perception data, translating and prioritizing sensory information for use by the strategic component. Additionally, the tactical layer employs specialized controllers, algorithms, and models to convert abstract commands from OntoAgent in the strategic layer into precise, actionable robot operations.

For example, a command such as \textsc{PICK(key, at-position1)}, from the OntoAgent to ``pick up a key on the ground," triggers a well-defined sequence of tasks and actions, including object identification, positional assessment, trajectory computation, and actuation, in the tactical layer. Furthermore, the tactical layer manages reactive behaviors, including collision avoidance, through dedicated controllers.

The tactical layer of HARMONIC is built around the Behavior Trees (BTs) \cite{colledanchise2018behavior}, which enable effective reactive control and support the system's safety and operational requirements in dynamic environments. The tactical layer may also include advanced control policies for whole-body compliant control \cite{iannotta2022heterogeneous, colledanchise2018behavior} with motion planning, path planning \cite{olsson2016behavior}, etc. BTs can also be used for representing skills and low-level plans that can also be shared between the robots \cite{venkata2023kt,oruganti2023ikt1}. 

\begin{figure}[h]
    \centering
    \includegraphics[width=\linewidth]{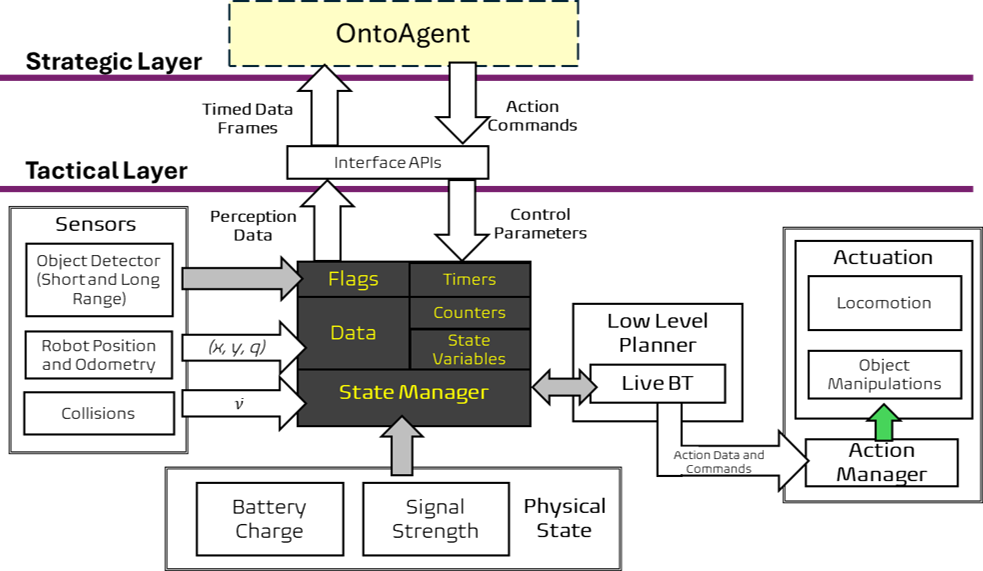}
    \caption{Data flow and control diagram showing the connections between the BTs in the tactical layer and the OntoAgent in the strategic layer in the HARMONIC System.}
    \label{fig:Harmonic-sys}
\vspace{-8pt}
\end{figure}

In the context of HARMONIC, the state manager module (a blackboard) keeps track of condition and state variables on the tactical layer \cite{colledanchise2021implementation}. As shown in Fig. \ref{fig:Harmonic-sys}, this allows for efficient querying and updating of the system's state during operation based on the sensory inputs and action commands received from the strategic layer. Sensory inputs are continuously written into the State Manager (SM), where the attention module filters and packages this perception data for the OntoAgent in the strategic layer. Simultaneously, the high-level action commands sent by the OntoAgent through the Interface APIs are also unpacked for updating the variables in the SM.

While sensor data and physical state tracking operate in real time, the strategic layer runs asynchronously. Real-time control is achieved through the data available in the SM, which is accessed in parallel by the LiveBT (a BT instance), a module that chooses policies and actions stored in the Action Manager libraries. This architecture enables HARMONIC to maintain real-time behavior while supporting higher-level planning capabilities.

These APIs update the placeholder variables within the State Manager, ensuring accurate and timely control of the tactical layer operations.

The HARMONIC architecture aligns with Kahneman's dual-system approach \cite{kahneman2011thinking}, with the OntoAgent in the strategic layer implementing System 2 (slow, deliberative reasoning) and the BTs in the tactical layer implementing System 1 (fast, intuitive responses). This structure allows for integrating reflexive and deliberative functioning, flexible scheduling and adaptation, which in turn enables real-time prioritization of goals and actions. As a result, HARMONIC effectively handles computational delays, contingencies, safety concerns, and resource optimization through a combination of low-level planning and reactive algorithms.








\section{Task Design}
\label{sec:plans}
We evaluated the HARMONIC framework's teaming strategy using a simulated search task set in an apartment environment, as illustrated in Fig. \ref{fig:Evaluation_Results}.6. The scenario involves an HMR team consisting of an unmanned ground vehicle (UGV), a drone, and a human team member. The objective is to find a set of lost keys. The human initiates the task by remotely communicating with the robotic team members in English. The simulation environment, developed in Unity, incorporates abstracted versions of robot behaviors. This approach allowed us to focus on key aspects of the HARMONIC framework's performance while simplifying the implementation of complex robot actions that are not central to our current work.

\subsection{Heterogenous Robot Team Configuration}
The simulation evaluation involved two functionally and structurally heterogeneous robots -- a UGV and a drone -- offering complementary capabilities and perspectives \cite{ov2020impact}. Each robot is equipped with sensors for localization, collision detection, and object recognition, facilitating coordinated and efficient operations.

The UGV's  (see Fig. \ref{fig:bts_robots}.a) can access ground-level areas and maneuver under obstacles and surfaces. The drone, on the other hand, complements the UGV by performing aerial search operations. The drone's (see Fig. \ref{fig:bts_robots}.b) vertical mobility and sensing capabilities allow it to scan environments extensively from above, navigating over obstacles and accessing areas difficult for the UGV to access. Since the robots communicate in natural language, their dialogue and actions are comprehensible to humans, and humans can intervene when necessary.



Both robots operate independent instances of HARMONIC, with different content parts (for example, there is no presumption that robots have identical ontological world models). For the experiment, the UGV acts as the team leader at the task level, responsible for managing interactions with the human operator and coordinating with the drone. The team leader assignment is presently arbitrary; however, future work aims to develop and evaluate heuristic methods for dynamic team hierarchies. This advancement will optimize leader selection based on contextual factors and the individual capabilities of each agent relevant to specific tasks.

\subsection{Development of Distributed Plans}

Plans are distributed between strategic and tactical components in both the UGV and the drone. The high-level plans (ontological script instances) are divided into steps communicated to the tactical layer as action commands, as described earlier.

\subsubsection{High-level planning}
The ontologies of agents in HARMONIC contain scripts for complex events that a particular agent can perform or understand other agents performing. Plans are instances of scripts with parameter values set for the situation at hand. These are instantiated by executing one or more of the ontologically stored meta-scripts, that is, scripts that HARMONIC robots instantiate to guide the generation of plans from scripts in their operational domain. One such meta-script is the \textsc{COLLABORATIVE-ACTIVITY} script, whose purpose is to help agents operating in teams organize themselves to accomplish a goal.

The details of the \textsc{COLLABORATIVE-ACTIVITY} script vary based on the agent’s role in the team. For example, the team (or task) leader’s script is typically more complex than the scripts of its subordinates since team leaders must select a domain script to instantiate into the plan to accomplish the target goal, resolve preconditions, and instruct their subordinates. Subordinates' scripts, by contrast, include awaiting instructions from the leader and passively absorbing any other related information that comes their way.

The leader’s script for collaboration is shown below. Since this is a meta-script, it can be applied to many specific scripts and starts with few parameter values populated. Part of its task is to fill in task-specific parameter values. Select comments are provided for this human-readable formalism. For further details on this formalism, please refer to \cite{english2020ontoagent,mcshane2021linguistics,mcshane2024agents}.

\begin{codeblock}
\textbf{@COLLABORATIVE-ACTIVITY (leader)}
[INIT]
  *identify-team-members
[SELECT-PLAN]
  \color{darkgreen}{// Select a plan from available options.}
  RUN *identify-candidate-plans
  RUN *select-plan
[PRECONDITIONS]
  \color{darkgreen}{// Preconditions will be determined once}
  \color{darkgreen}{// a plan has been picked.}
[SUGGEST-PLAN] 
  \color{darkgreen}{// Propose the selected plan to the team.}
  RUN NEW @PROPOSE-PLAN
[RUN-PLAN]
  \color{darkgreen}{// The leader hasn’t picked a plan yet.}
\end{codeblock}

The subordinate’s script for collaboration is:

\begin{codeblock}
\textbf{@COLLABORATIVE-ACTIVITY (subordinate)}
[INIT]
  *identify-team-members
[WAIT-FOR-PLAN]
  \color{darkgreen}{// Wait for a plan from the leader before }
  \color{darkgreen}{// continuing.}
  AWAIT \$.HAS-COLLABORATIVE-PLAN ISA @EVENT
[RUN-PLAN]
  \color{darkgreen}{// The subordinate has no plan yet.}
\end{codeblock}

Having instantiated an instance of COLLABORATIVE-ACTIVITY as a meta-plan,  the leader will select the domain script to pursue and instantiate as a domain plan. In the example below, the selected script is \textsc{SEARCH-FOR-LOST-OBJECT}.  The details of the plan instantiated from the selected script can be placed into the \textsc{[RUN-PLAN]} section of \textsc{COLLABORATIVE-ACTIVITY}, and any preconditions that must be resolved can be added to the \textsc{[PRECONDITIONS]} section. An excerpt from the leader's script for \textsc{COLLABORATIVE-ACTIVITY} is shown below.

\begin{codeblock}
\textbf{@COLLABORATIVE-ACTIVITY (leader)}
[INIT]
  *identify-team-members
[SELECT-PLAN]
  \color{darkgreen}{// Select a plan from available options.}
  RUN *identify-candidate-plans
  RUN *select-plan
[PRECONDITIONS]
  \color{darkgreen}{// Preconditions from the selected plan }
  \color{darkgreen}{// to be checked.}
  RUN NEW @REQUEST-OBJECT-TYPE
  RUN NEW @REQUEST-OBJECT-FEATURES
  RUN NEW @REQUEST-LAST-SEEN-AT
  RUN NEW @REQUEST-LOCATION-CONSTRAINED
[SUGGEST-PLAN] 
  \color{darkgreen}{// Propose the selected plan to the team.}
  RUN NEW @PROPOSE-PLAN
[RUN-PLAN]
  \color{darkgreen}{// Execute the selected plan, step by step.}
  RUN NEW @SEARCH-FOR-LOST-OBJECT
\textbf{@SEARCH-FOR-LOST-OBJECT}
  [SEARCH-ZONES]
    \color{darkgreen}{// Search in each ZONE the agent knows.}
    \color{darkgreen}{// Stop searching when the object has been located.}
    \color{darkgreen}{// Report the results (if necessary) to the team.}
    FOR #ZONE-1 IN #LOCATION-1.SEARCHABLE-ZONE 
      RUN ASYNC AWAIT *search 
      INTERRUPT WHEN #OBJECT-1.LOCATION KNOWN
      RUN *consider-reporting
\end{codeblock}

Once the leader has worked through the preconditions, it can move to \textsc{SUGGEST-PLAN}, in which it will share the domain plan with subordinates through natural language dialog.  These subordinates will receive this information and adopt their assignment(s) within the suggested plan, resulting in the team adopting a shared collaborative plan.

 \subsubsection{Low-level control}
 For low-level planning and control on the tactical layer, we use Behavior Trees (BTs). 

 BTs provide a flexible and intuitive way to design control actions and define task-planning hierarchies for robots \cite{colledanchise2018behavior}.  Their modular design enables easy modification and scaling of behaviors, making BTs particularly suitable for complex hierarchical task planning, in the tactical layer. 
 
\begin{figure}[h]
     \centering
     \includegraphics[width=\linewidth]{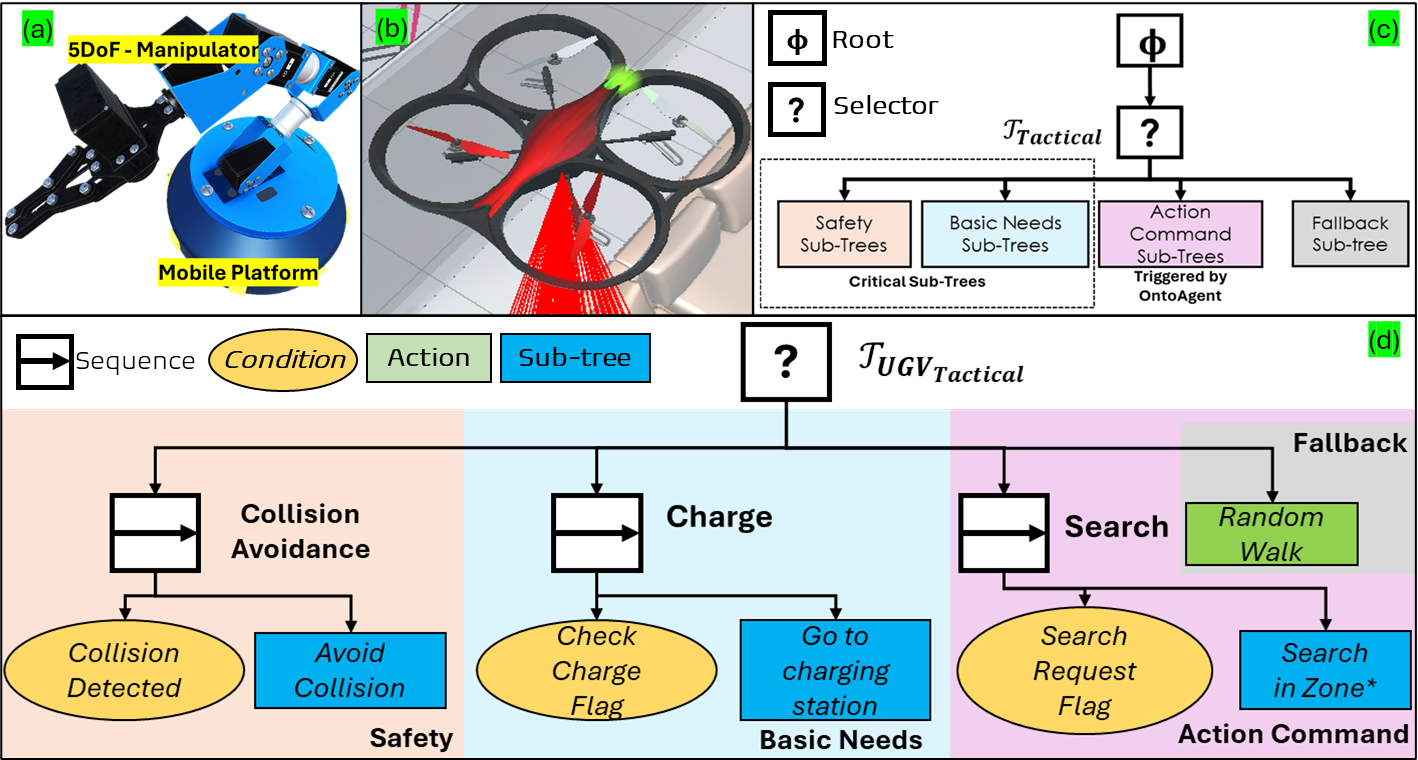}
     \caption{\textbf{(a)} UGV for ground-level exploration. \textbf{(b)} Parrot drone conducting aerial scan. \textbf{(c)} BT design template for the tactical layer of robots. \textbf{(d)} Sample BT on UGV.}
     \label{fig:bts_robots}
     \vspace{-8pt}
 \end{figure}

 BTs for both robots follow a standardized design template prioritizing safety and operational needs as shown in Fig. \ref{fig:bts_robots}.c. This design leverages the conventional left-to-right execution priority, ensuring the robot's critical needs and priorities. Collision avoidance subtrees are positioned leftmost, followed by robot-needs sub-trees, then action-command sub-trees. On the far right, fallback sub-trees handle behaviors like random exploration. An excerpt from the UGV's BT is shown in Fig. \ref{fig:bts_robots}.d.


 \begin{figure*}[!ht]
    \centering
    \includegraphics[width=\linewidth]{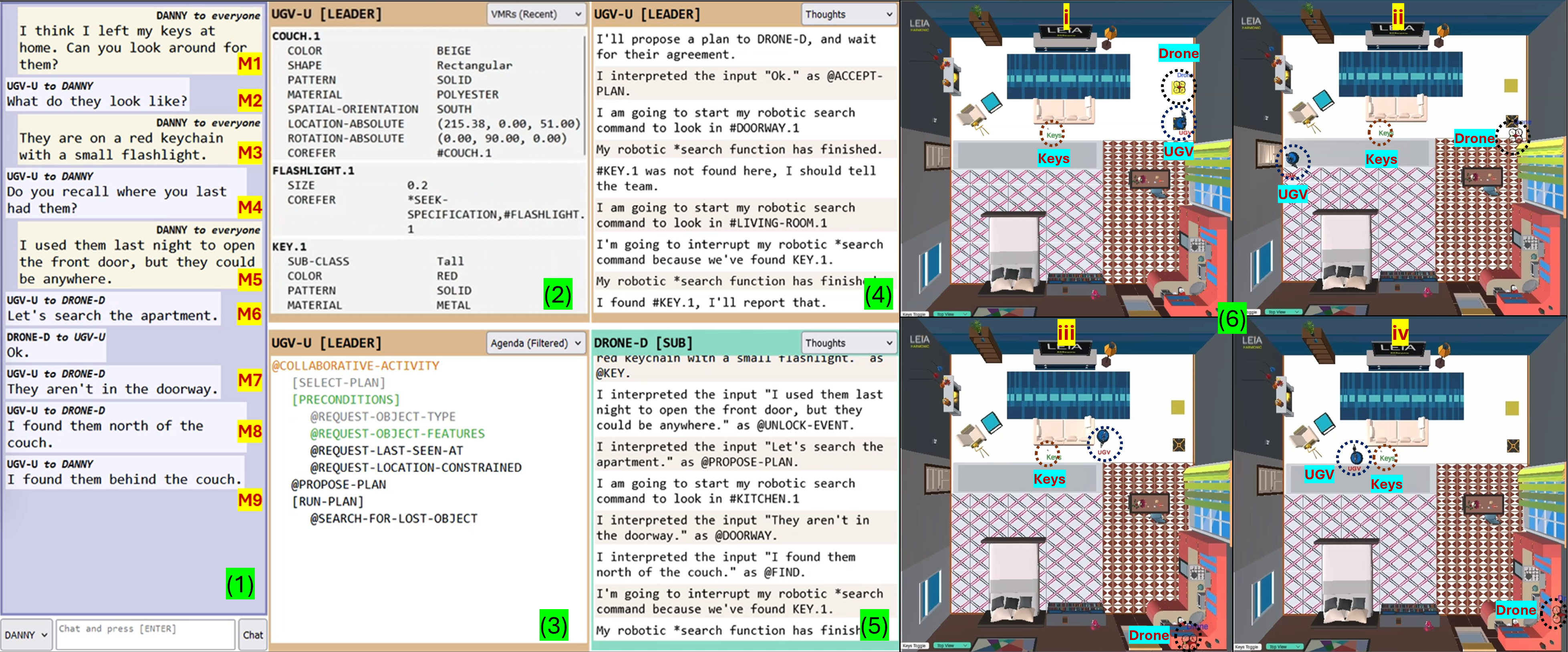}
    \caption{\textbf{(1)} Chat transcript between Danny (human) and the robots. \textbf{(2)} UGV's (leader) sample Vision Meaning Representations (VMRs) of detected objects by the robots \textbf{(3)} Leader's agenda. \textbf{(4, 5)} Complete thought transcripts of UGV and Drone. \textbf{(6)} Multiple instances from the simulated apartment environment showing the UGV, the drone, and the keys. {\textbf{(i)}} Robots at their base stations. {\textbf{(ii, iii)}} Robots searching the apartment. {\textbf{(iv)}} UGV finds the keys.}
    \label{fig:Evaluation_Results}
    \vspace{-10pt}
\end{figure*}
\section{Evaluation Results}
\label{sec:evaluation}
 The simulation environment features an integrated chat UI through which the human operator (Danny) interacts with the robots. As shown in Fig. \ref{fig:Evaluation_Results}(1-5), the interface comprises multiple components: a central chat window, visualization widgets for the robots' language interpretations and visual perceptions (VMRs), natural language representations of robot cognition, and structured displays of their goals and plans. This design offers transparency into the robots' decision-making processes and cognitive operations. The analysis below refers to specific labeled elements in Fig. \ref{fig:Evaluation_Results}. Please view the accompanying video for a better understanding.


\subsubsection{Task Initiation}
The scenario begins with Danny initiating the task by sending the robots a message \textbf{(M1)}. This input triggers the team leader (UGV) to place a \textsc{COLLABORATIVE-ACTIVITY} on its agenda (Fig. \ref{fig:Evaluation_Results}.3) and launch the \textsc{SEARCH-FOR-LOST-OBJECT} plan. Concurrently, the drone awaits instructions.

\subsubsection{Information Gathering}
The UGV proceeds to verify preconditions for the \textsc{SEARCH-FOR-LOST-OBJECT} plan (Fig. \ref{fig:Evaluation_Results}.3). Although the object type is already known, information about its features and last-seen location would be useful. The UGV queries Danny for this information \textbf{(M2/M4)}, and receives responses (M3/M5). The latter response prompts the system to prioritize \textit{entry-way} in the search sequence \textbf{(ii)}.

\subsubsection{Search Execution}
With preconditions met, the UGV instructs the drone to initiate the search \textbf{(M6)}, and both robots begin exploring their assigned areas as shown in \textbf{(ii-iv)}. The individual tactical modules on the robots control the search process using a waypoint strategy, while the strategic (cognitive) module maintains awareness of area existence without directly guiding robot navigation. This approach allows for efficient local path planning while preserving high-level planning in the strategic layer.

\subsubsection{Communication and Coordination}
Throughout the search, robots report their findings to each other and the human \textbf{(M7-8, ii-iv)}. When a robot fails to locate the keys in a searched area, it communicates this to its partner.

\subsubsection{Task Completion}
The VMRs widget for UGV (Fig. \ref{fig:Evaluation_Results}.2) displays the object detection results that the strategic layer processes from sensing frames communicated by the tactical layer.
 During the search execution, the strategic module continuously analyzes the sensor data frames, grounding these VMRs against the instance of the KEY object stored in its episodic memory. When the keys are found \textbf{\textbf(iv)} -- the features match, the search is halted by either of the robots (leader in this case), informs the team \textbf{(M8)}, and reports to Danny \textbf{(M9)}. Notably, the UGV uses different language constructs when communicating with the drone versus Danny in \textbf{(M9)}, demonstrating the cognitive agent's ability to generate context-appropriate language.

\section{Conclusion and Future Work}
This work presents a step towards the development of cognitive robots capable of effectively collaborating in heterogeneous human-robot teams. The HARMONIC framework, with its dual-layer architecture that combines strategic cognitive processing and tactical control, allows robots to exhibit metacognitive abilities, including team task understanding, episodic memory utilization, and multimodal perception interpretation. By integrating natural language communication and reasoning capabilities, our approach facilitates complex team interactions, enhances explainability, and builds trust, the key factors in human-robot collaboration. 


Our demonstration system of a simulated search task involving two robots and a human teammate illustrates the potential for HARMONIC to be applied to MRS problems. Future research will focus on expanding the range of tasks and scenarios in which HARMONIC agents can successfully operate; enhancing the robots' learning capabilities; conducting real-world trials using physical robots to validate the framework's effectiveness in diverse operational contexts; and developing collaborative online planning strategies in dynamic environments;

\ifblind
\else
\section*{ACKNOWLEDGMENT}
This work was supported in part by ONR Grant \#N00014-23-1-2060. The views expressed are those of the authors and do not necessarily reflect those of the Office of Naval Research.
\fi

\bibliographystyle{IEEEtran}


\end{document}